\documentclass[10pt,twocolumn,letterpaper]{article}

\usepackage[pagenumbers]{cvpr} 

\usepackage{graphicx}
\usepackage{amsmath}
\usepackage{amssymb}
\usepackage{booktabs}

\usepackage[pagebackref,breaklinks,colorlinks]{hyperref}

\usepackage[capitalize]{cleveref}
\crefname{section}{Sec.}{Secs.}
\Crefname{section}{Section}{Sections}
\Crefname{table}{Table}{Tables}
\crefname{table}{Tab.}{Tabs.}

\def\confName{CVPR}
\def\confYear{2022}

\begin{document}

\title{Dual-Stream Transformer for Generic Event Boundary Captioning}

\author{Xin Gu$^{1}$, Hanhua Ye$^{1}$, Guang Chen$^{1}$, Yufei Wang$^{1}$, Libo Zhang$^{2}$, Longyin Wen$^{1}$ \\
$^{1}$ByteDance Inc., Mountain View, USA. \\
$^{2}$Institute of Software Chinese Academy of Sciences, Beijing, China. \\
{\tt\small \{guxin.6, yehanhua, guang.chen, yufei.wang, longyin.wen\}@bytedance.com}, \\ {\tt\small libo@iscas.ac.cn}}

\maketitle

\begin{abstract}
    This paper describes our champion solution for the CVPR2022 Generic Event Boundary Captioning (GEBC) competition.
    GEBC requires the captioning model to have a comprehension of instantaneous status changes around the given video boundary, 
    which makes it much more challenging than conventional video captioning task.
    In this paper, a Dual-Stream Transformer with improvements on both video content encoding and captions generation is proposed:
    (1) We utilize three pre-trained models to extract the video features from different granularities. 
    Moreover, we exploit the types of boundary as hints to help the model generate captions.
    (2) We particularly design an model, termed as Dual-Stream Transformer, to learn discriminative representations for boundary captioning.
    (3) Towards generating content-relevant and human-like captions, we improve the description quality by designing a word-level ensemble strategy.
    The promising results on the GEBC test split demonstrate the efficacy of our proposed model.
    Code is available at \url{https://github.com/GX77/Dual-Stream-Transformer-for-Generic-Event-Boundary-Captioning}.
    
\end{abstract}

\section{Introduction}
\label{sec:intro}
Introduced by~\cite{GEBC}, Generic Event Boundary Captioning (GEBC) is an advanced multi-modal task which seeks to automatically generates descriptions for the subject, status before and status after of the given video boundary.
The conventional video captioning task aims to describe the overall event happening at the given video clip.
In contrast, GEBC is much more challenging in that it requires the comprehension of instantaneous and fine-grained status changes inside the video.
Therefore, exploiting multi-modal video features of various granularity and learning discriminative representations are essential for current task.

Since ``pre-training and fine-tuning" has become a de facto paradigm in many vision-language tasks, such as video captioning~\cite{CLIP4Caption,swinbert} and video retrieval~\cite{videoclip}, 
we exploit pre-trained CLIP~\cite{CLIP} and VideoSwin~\cite{videoswin} as backbones to extract video appearance features and video motion features, and fine-tune them during training stage.
Meanwhile, Faster-RCNN~\cite{RCNN} is utilized to extract region of interest of given videos.
Additionally, we utilize the ``types of boundary" labels as the language-modality input to help the model generate more accurate descriptions for boundaries.

In order to learn discriminative representations for video boundaries, the extracted multi-modal features are input into our especially designed Dual-Stream Transformer. 
The local stream focuses on the crucial components in the multi-source features, \textit{e.g.}, captions, the appearance of detected object regions, and boundary types. 
The global stream focuses on learning the interactions between the multi-source features, \textit{e.g.}, captions, the appearance of video frames, video motions, and boundary types.
Moreover, we design a word-level ensemble strategy during inference stage to generate more reasonable results and improve description quality.

Our whole model is trained in an end-to-end fashion and no additional data is used during training process. 
The proposed Dual-Stream Transformer solution finally won the championship on the GEBC~\cite{GEBC} competition.

\begin{figure*}[htb]
 \centering
 \includegraphics[ width=0.9\linewidth]{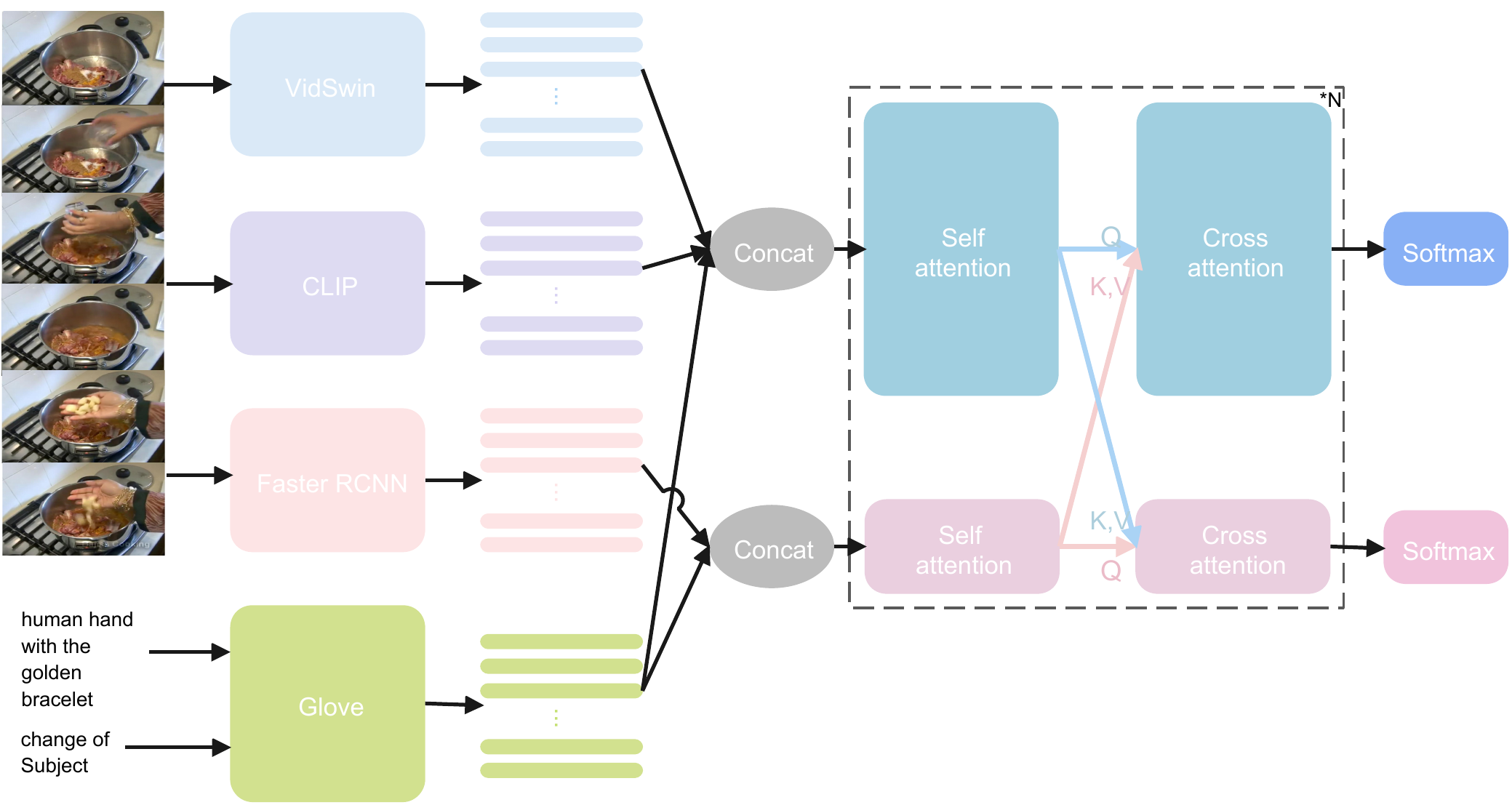}
 \caption{The proposed model consists of several backbones and the dual-stream transformer. 
 The VideoSwin and CLIP extract video appearance and motion features respectively, and are fine-tuned during training.
 The Faster-RCNN extracts object regions, providing detailed information for later representation learning.
 Finally, these extracted visual features and GloVe-embedded ``types of boundary" labels, along with generated captions, are input into the dual-stream transformer to learn representations,
 generating descriptions for given video boundaries.}
 \label{fig1}
\end{figure*}
\section{Method}
Figure~\ref{fig1} demonstrates the architecture of our model. 
First, the pre-trained VideoSwin and CLIP are employed to extract video motion and appearance features respectively. 
To fully exploit the video information, we embed ``types of boundary" labels with pre-trained GloVe~\cite{glove} as hints to help the model generate accurate captions for boundaries.
Next, the extracted visual features and linguistic features are input into the dual stream transformer to construct discriminative video representations.
During description generation, the appearance features, motion features, region features, and boundary type embeddings, along with generated captions, are fed into the dual-stream transformer.
For ensemble, we train multiple models with different settings and ensemble their outputs on the word-level, gaining final strong results.
Details will be elaborated as follows.

\subsection{Multi-modal features exploiting}
Fully exploiting multi-modal video features is vital for our model to caption generic event boundaries. 
Features of visual modality, including appearance, motion, and region features, are essential for the model to understand the overall events and instantaneous changes of the videos.
On the other hand, features of linguistic modality, including boundary types and generated descriptions, are crucial for descriptions' fluency and accuracy.

CLIP was pre-trained on 400 million image-text pairs~\cite{400m}, which making it learn adequate visual representations since it can take advantage of the information from both visual and linguistic modality.
The architectures of CLIP's image encoder are various. Here we choose ViT-B/32 as our appearance features extractor.
Benefiting from multi-head self-attention and 3D shifted windows, VideoSwin\cite{videoswin}, which is pre-trained on Kinetics400\cite{kinetics}, is perfect for action recognition. Here we choose Swin-T based model as our motion features extractor, which is pre-trained on the ImageNet-1K\cite{1k}.
During training, we fine-tune these two backbones along with our designed dual-stream transformer to obtain better video features.

In order to obtain detailed video information, we exploit ResNet152-based Faster-RCNN~\cite{RCNN} as the object detector, which is pretrained on the Visual Genome dataset~\cite{vg}.
Given a sequence of video frames, we sample keyframes at 2FPS. 
For each keyframes, we select the Top-10 regions with the highest confidence. 

Boundary types, such as ``change of action" and ``change of subject", can inform our captioning model the type of instantaneous changes around the boundary, which may assist the model to generate more accurate descriptions.
Meanwhile, the generated captions can provide linguistic context for our model, which is helpful for maintaining the description's fluency.
We embed both of the boundary types and generated captions as our linguistic features by using the GloVe~\cite{glove} that has been pre-trained on the Wikipedia and Gigaword 5 Data.

\subsection{Dual-Stream Transformer}
For full interaction of the different modal information, we designed a two-branch network structure that contains both global and local branches.
The input to the global stream is appearance feature~($f_A$), motion feature~($f_M$), boundary type embedding~($e_B$) and captioning embedding~($e_C$), while the input to the local stream is region feature~($f_R$), boundary type embedding~($e_B$) and captioning embedding~($e_C$).
Firstly, the different feature information is first encoded uniformly into feature vectors of the same dimension by the linear layer.
\begin{align}
& X_{A} = \text{Linear}_A(f_{A}) \notag \\ 
& X_{M} = \text{Linear}_M(f_{M}) \notag \\ 
& X_{R} = \text{Linear}_R(f_{R}) \notag \\ 
& X_{B} = \text{Linear}_B(e_{B}) \notag \\ 
& X_{C} = \text{Linear}(e_{C})           
\end{align}
Then, the different feature vectors belonging to the same stream are concatenated together and fed into the self attention modules.
\begin{align}
& X_{l}  = [X_{R}, X_{B}, X_{C}] \notag \\
& X_{g} = [X_{A}, X_{M}, X_{B}, X_{C}] \\
& Q_{l}, K_{l}, V_{l} = X_{l}W^{Q_{l}}, X_{l}W^{K_{l}}, X_{l}W^{V_{l}} \notag \\
& Q_{g}, K_{g}, V_{g} = X_{g}W^{Q_{g}}, X_{g}W^{K_{g}}, X_{g}W^{V_{g}} \\
& X_{l}^{'} = \text{SelfAttention}(Q_{l},K_{l},V_{l}) \notag \\
& X_{g}^{'} = \text{SelfAttention}(Q_{g},K_{g},V_{g})
\end{align}
where $W$ is the learnable parameters. Next, the model aligns the information from the two streams with features by using the cross attention module.
\begin{align}
& Q_{l}^{'}, K_{l}^{'}, V_{l}^{'}  = X_{l}^{'}W^{Q_{l}}, X_{g}^{'}W^{K_{l}}, X_{g}^{'}W^{V_{l}} \notag \\
& Q_{g}^{'}, K_{g}^{'}, V_{g}^{'} = X_{g}^{'}W^{Q_{g}}, X_{l}^{'}W^{K_{g}}, X_{l}^{'}W^{V_{g}} \\
& X_{l}^{''}  = \text{CrossAttention}(Q_{l}^{'},K_{l}^{'},V_{l}^{'}) \notag \\
& X_{g}^{''} = \text{CrossAttention}(Q_{g}^{'},K_{g}^{'},V_{g}^{'})
\end{align}
where $W$ is the learnable parameters. Finally, the output of the two streams will be fed through the softmax layer to get the predicted words.

\subsection{Ensemble Strategy}
In order to achieve a more reasonable caption result, we ensemble 3 models with different initialization parameters.These models are all independent of each other in generating the predicted words for the current moment.At moment $t$, when each model generates a word probability vector for the current moment. We first sum all the probability vectors, and the word with the highest probability is used as the predicted word at moment $t$. This word is then used as the input word for each model at moment $t+1$.

\subsection{Training}
We use the cross entropy loss to guide the training of our method. The loss is formally given by
\begin{equation}
\begin{split}
{\cal L}=-\sum_{t=1}^T\big(\lambda_1\log(p_{\theta_{l}}(y_t^*|y^*_{1:t-1},f_{A},f_{M},e_{B}))+ \\ \lambda_2\log(p_{\theta_{g}}(y_t^*|y^*_{1:t-1},f_{R},e_{B}))\big)
\end{split}
\end{equation}
where $\lambda_1$ and $\lambda_2$ are the parameters that balance the two streams, $T$ is the length of the caption and $y^*_{1:t-1}$ is the ground truth. Our model is trained in an end-to-end manner, and use no additional data during training process.

\section{Experiments}
\subsection{Implementation Details} 
\textbf{Datasets.}
We validate our method on the Kinetic-GEBC dataset~\cite{GEBC} and it includes 176,681 boundaries in 12,434 videos selected from all categories in Kinetic-400~\cite{kinetics}. Each annotation consists of several boundaries inside a video. Each video contains 1 to 8 annotations from different annotators, where the boundaries' location are not the same.
Following \cite{GEBC}, we use $8,269$ videos for training, $2,082$ videos for validation, and $2,082$ videos for testing, respectively.

\textbf{Evaluation metric.}
Following \cite{GEBC}, we employ three standard metrics to evaluate our method, including  
CIDEr~\cite{CIDEr} for tf-idf weighted n-gram similarity, SPICE~\cite{SPICE} for similarity between the scene graphs of the text and ROUGE\_L~\cite{ROUGE} for n-gram recall. Meanwhile, we separate the prediction captions into Subject, Status Before and Status After, and then compute the similarity score of each item with the ground truth. Finally, the average score is computed across three items for each metric.

\textbf{Network architecture.}
Our model sample 2 frames per second in each video and use the CLIP~\cite{CLIP}, VideoSwin~\cite{videoswin} and Faster RCNN~\cite{RCNN} as the feature extractors. Our model employs a shared encoder-decoder network and per layer contains a self attention module and a cross attention module. The number of layers in the model is set to $3$, the dimension $d$ of each layer is set to $768$ and the number of heads is set to $12$. 

\textbf{Training.} 
We train our model in an end-to-end way and all the experiments are conducted on a machine with 8 Tesla V100 GPUs. We use the Adam optimizer with the initial learning rate of $1e-4$. The proposed method is trained at most $10$ epochs with the batch size $100$. The hyper-parameters $\lambda_1$ and $\lambda_2$ are set to $0.5$, and $0.5$, empirically.

\subsection{Ablation Study}
To verify the role of the different modules, we perform the ablation experiments, as shown in \ref{Table 1}. For each input video, our feature extractor transforms it into three features, including appearance feature by CLIP~\cite{CLIP}, motion feature by VideoSwin~\cite{videoswin} and region feature by Faster RCNN~\cite{RCNN}, each of which improves the performance of the model. Meanwhile, in order to make full use of the information of the video, our proposed model also considers the boundary type information of the video. This information can help the model to distinguish different states (Subject, Status Before and Status After) when generating descriptions, which improves the AVG score by 14.3\%. Finally, to enhance the robustness of the model, we ensemble three models with different initialization parameters, which also improves the performance of the model.

\begin{table}[!ht]
\centering
\setlength{\tabcolsep}{3.5pt}
\begin{tabular}{ccccc|c}
\hline
Region  &  Motion  & Appearance & Video & & \\
Feature &  Feature &  Feature & Type  & Ensemble & AVG  \\ \hline
\checkmark & \checkmark  & ~ & ~ & ~ & 63.34  \\ 
\checkmark  & \checkmark  & \checkmark  & ~ & ~ & 64.21  \\ 
\checkmark  & \checkmark  & \checkmark  & \checkmark  & ~ & 73.36    \\
\checkmark  & \checkmark  & \checkmark  & \checkmark  & \checkmark  & 74.30  \\ \hline
\end{tabular}
\caption{Ablation study on the Kinetic-GEBC dataset val set.}
\label{Table 1}
\end{table}

\subsection{Overall Results}
As shown in Table \ref{Table 2}, our proposed method achieve 74.39 average score on the Kinetic-GEBC test set. Specifically, our method improves $98.3$\% absolute CIDEr score compared to the baseline. The experimental results show that our proposed method can perform the generic event boundary captioning task well.

\begin{table}[!ht]
\centering
\setlength{\tabcolsep}{8.0pt}
\begin{tabular}{ccccc}
\hline
Model & AVG & SPICE & ROUGE\_L & CIDEr  \\ \hline
Baseline & 40.80 & 19.52 & 28.15 & 74.71  \\ \hline
Ours & 74.39 & 33.86 & 41.19 & 148.13  \\ \hline
\end{tabular}
\caption{The final results on the Kinetic-GEBC dataset test set.}
\label{Table 2}
\end{table}

\subsection{Qualitative Results}
\begin{figure}[]
 \centering
 \includegraphics[ width=1\linewidth]{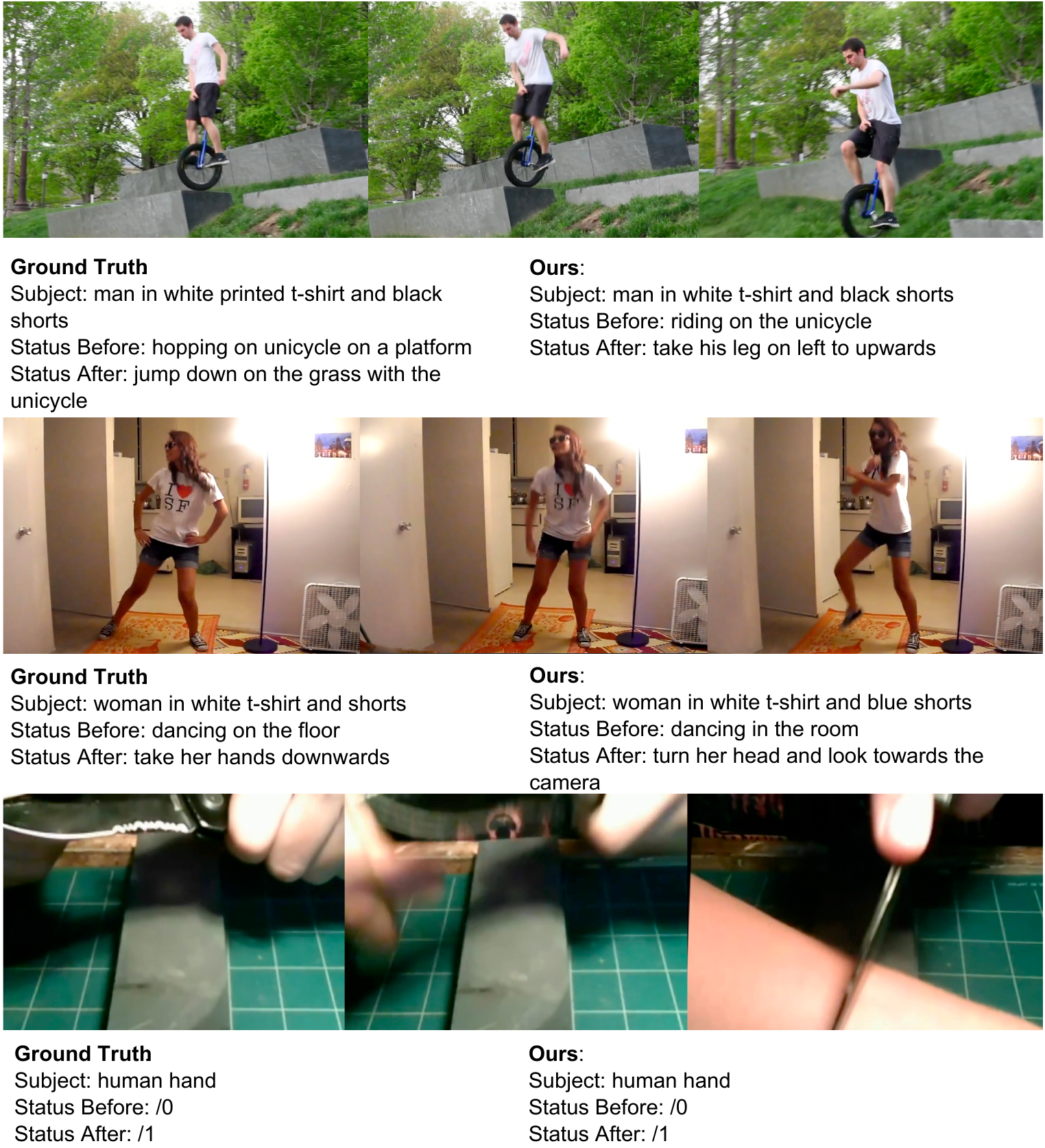}
 \caption{Exemplar on the Kinetic-GEBC dataset.}
 \label{fig2}
\end{figure}

We show some qualitative results in Figure \ref{fig2}. For each video boundary, three types of descriptions are generated, including subject, status before and status after. As shown in the figure, our model can generate high-quality captions.

\section{Conclusion}
In this work, we propose a Generic Event Boundary Captioning model, termed as Dual-Stream Transformer,
which improve the performance by exploiting multi-modal video features, learning discriminative representations and designing a word-level ensemble strategy.
Experimental results indicate that each sub-module or strategy contribute to the captioning metrics.
Finally, the proposed method achieves the best results in the GEBC competition.

{\small
\bibliographystyle{ieee_fullname}
\bibliography{egbib}
}

\end{document}